%% file: root.tex
\documentclass[letterpaper, 10 pt, conference]{ieeeconf}  

\IEEEoverridecommandlockouts                              

\input{tex/preamble}

\title{\LARGE \bf
BEVIO: Efficient Bird’s-Eye-View based Sparse-Update Visual-Inertial Odometry for Lunar Day-Night Navigation}

\author{Mohit Singh$^{1,2,*}$, Shehryar Khattak$^{1}$, Ashish Goel$^{1}$, Michael Paton$^{1}$, Kostas Alexis$^{2}$, and Issa A. Nesnas$^{1}$
\thanks{$^{*}$Corresponding author: {\tt\small mohit.singh@ntnu.no}}
\thanks{$^{1}$Jet Propulsion Laboratory, California Institute of Technology, Pasadena, CA, USA.}
\thanks{$^{2}$Autonomous Robots Lab at the Norwegian University of Science and Technology, Trondheim, Norway}
\thanks{The research was carried out at the Jet Propulsion Laboratory, California Institute of Technology, under a contract with the National Aeronautics and Space Administration (80NM0018D0004).}
\thanks{\copyright 2025. All rights reserved.}
}

\begin{document}

\maketitle
\thispagestyle{empty}
\pagestyle{empty}

\begin{abstract}
Visual–Inertial Odometry (VIO) provides smooth, high-rate state estimates and has been widely used for robotic navigation in both terrestrial and planetary applications. However, its performance is typically dependent on the frequency of visual updates, which is a challenge for planetary rovers operating under extreme resource constraints and low frame rates. This work investigates enabling reliable VIO with very sparse visual updates for lunar rover applications, addressing both day and night-time operations where feature associations become especially difficult under self-illumination conditions. We propose a Bird’s Eye View (BEV)–based image matching scheme that remains robust to larger inter-frame motions and more reliable feature matching despite significant visual appearance changes. We extensively evaluate our proposed approach, BEVIO, through high-fidelity photorealistic lunar and real-time robotic experiments conducted using a half-scale lunar rover, in a long-term day–night deployment at Plaster City, CA, USA. The results demonstrate that our method enables reliable day and nighttime self-illuminated traverses at visual update rates as low as 0.25 Hz, underscoring its suitability for navigation on power- and compute-limited lunar rovers.

\end{abstract}

\section{INTRODUCTION}
\input{tex/1_Introduction}

\section{RELATED WORK}
\input{tex/2_RelatedWork}

\section{METHOD}
\label{sec: bev}
\input{tex/3_BEV}

\section{INTEGRATION INTO VISUAL-INERTIAL ODOMETRY}
\label{sec: vio}
\input{tex/4_VIOIntegration}

\section{EXPERIMENTS}
\input{tex/5_experiments}

\section{CONCLUSIONS}
\input{tex/6_Conclusion}


\addtolength{\textheight}{-12cm}   

\bibliographystyle{IEEEtran}
\bibliography{BIB/main}

\end{document}

%% file: tex/preamble.tex
\usepackage{graphics} 
\usepackage{graphicx}

\usepackage{subcaption} 

\usepackage{epsfig} 
\usepackage{amsmath} 
\usepackage{amssymb}  

\usepackage{bm}
\usepackage{mathtools}

\usepackage{algorithm}
\usepackage[noend]{algpseudocode} 
\algnewcommand\AAND{\textbf{ and }}
\algnewcommand\Or{\textbf{ or }}

\usepackage{color}
\usepackage{citesort}
\usepackage{flushend}
\usepackage{url}
\usepackage[breaklinks]{hyperref}
\usepackage[capitalise]{cleveref}
\usepackage{booktabs}
\usepackage{makecell}
\usepackage{multirow}
\usepackage{siunitx}
\usepackage{svg}
\usepackage[nolist,nohyperlinks]{acronym}
\input{tex/acronyms.tex}


\usepackage{tikz}
\usetikzlibrary{positioning, shapes.geometric, arrows.meta, decorations.pathreplacing, calligraphy}

\DeclareMathAlphabet{\pazocal}{OMS}{zplm}{m}{n}

\DeclareMathAlphabet{\mathpzc}{OT1}{pzc}{m}{it}

\usepackage{array}
\newcolumntype{C}[1]{>{\centering\arraybackslash}p{#1}}
\newcolumntype{M}[1]{>{\raggedright\arraybackslash}p{#1}}

\usepackage{array} 
\newcolumntype{L}[1]{>{\raggedright\let\newline\\\arraybackslash\hspace{0pt}}m{#1}}	
\newcolumntype{S}[1]{>{\centering\let\newline\\\arraybackslash\hspace{0pt}}m{#1}}
\newcolumntype{R}[1]{>{\raggedleft\let\newline\\\arraybackslash\hspace{0pt}}m{#1}}

\makeatletter
\renewcommand*{\@opargbegintheorem}[3]{\trivlist
  \item[\hskip \labelsep{\itshape #1\ #2}] \textit{(#3)}\ }
\makeatother

%% file: tex/acronyms.tex
\acrodef{method}[AOM]{ACRONYM OF METHOD}
\acrodef{gnss}[GNSS]{Global Navigation Satellite System}
\acrodef{ransac}[RANSAC]{Random Sample Consensus}
\acrodef{slam}[SLAM]{Simultaneous Localization And Mapping}
\acrodef{pca}[PCA]{Principal Component Analysis}
\acrodef{ekf}[EKF]{Extended Kalman Filter}
\acrodef{rmse}[RMSE]{Root Mean Square Error} 
\acrodef{ape}[APE]{Absolute Pose Error}
\acrodef{cfar}[CFAR]{Constant False Alarm Rate}
\acrodef{snr}[SNR]{Signal to Noise Ratio}
\acrodef{rcs}[RCS]{Radar Cross Section}
\acrodef{imu}[IMU]{Inertial Measurement Unit}
\acrodef{sgm}[SGM]{Segmi-Global Matching}
\acrodef{dnn}[DNN]{Deep Neural Network}
\acrodef{gru}[GRU]{Gated Recurrent Unit}
\acrodef{hpr}[HPR]{Hidden Point Removal}
\acrodef{raft}[RAFT]{Recurrent All-Pairs Field Transforms}
\acrodef{fov}[FOV]{Field of View}
\acrodef{mclab}[MC-lab]{Marine Cybernetics laboratory}
\acrodef{vio}[VIO]{Visual-Inertial Odometry}
\acrodef{vo}[VO]{Visual Odometry}
\acrodef{rcm}[RCM]{Refractive Camera Model}
\acrodef{sfm}[SFM]{Structure from Motion}
\acrodef{bev}[BEV]{Bird's-Eye View}
\acrodef{is}[IS]{Image Space}
\acrodef{fps}[FPS]{Frames Per Second}
\acrodef{rpe}[RPE]{Relative Position Error}
\acrodef{gpu}[GPU]{Graphics Processing Unit}

%% file: tex/1_Introduction.tex
Over the past five decades, robotic space explorers have enabled humankind to develop a rich understanding of our solar system. Future mission concepts continue to push the boundaries of exploration.  One such concept is the Endurance mission~\cite{keaneEnduranceLunarSouth,endurance2025missionProgress}, which plans to explore the South Pole-Aitken (SPA) Basin of Earth's Moon and traverse over 2000 km across four years. The Radioisotope Thermoelectric Generator (RTG) powered version of the rover targets a maximum speed of 1 km/hour ($\approx$\SI{0.28}{m/s}) and an ability to traverse during both the lunar day and night. The solar-powered version can only traverse during the day due to a lack of energy for anything but surviving the extreme cold of the lunar nights. This mobility represents a substantial increase over past missions, such as Mars 2020 Perseverance, which traversed at a top speed of \SI{0.042}{m/s}. Endurance is being designed to operate at higher traversal speeds with a higher degree of autonomy. Additional challenges arise in dark regions during nighttime missions or during daytime in dark shadows (e.g., from craters), where the rover needs to self-illuminate the environment in its proximity for sustained perception. Such a rover is limited by both onboard computing and energy; to this end, it is important to develop and study methods that can enable long-term, reliable performance while minimizing the computing and energy resources.
\begin{figure}[]
    \centering    \includegraphics[width=0.99\linewidth]{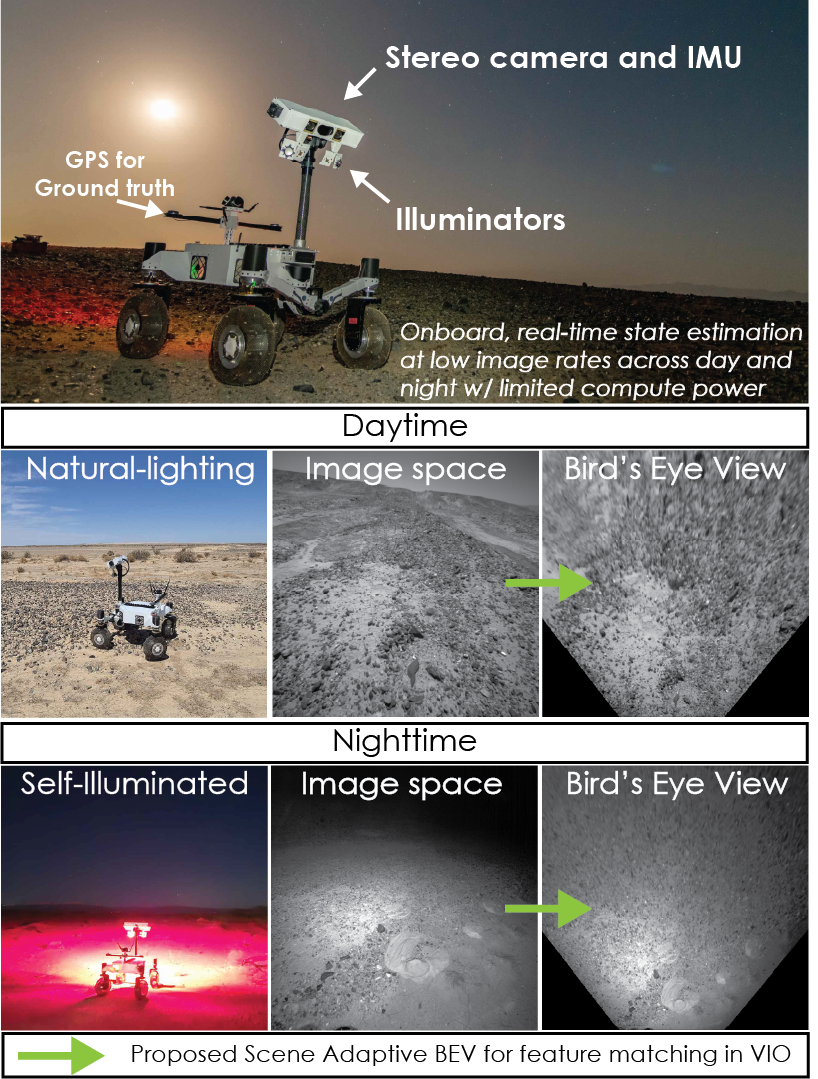}    \caption{\small Half-scale ERNEST rover, deployed in planetary-like desert terrain, navigating with onboard real-time state estimation. The challenges include fast traversal in day and nighttime operation, using constrained compute resources. Proposed solution employs a scene adaptive Bird's Eye View perspective to match features, enabling Visual-Inertial Odometry at lower image frame rates than the baseline approach, thus reducing the computational requirements.}
    \label{fig:ernest}
\vspace{-8px}
\end{figure}

\par We are investigating state estimation for the Endurance by utilizing a half-scaled rover ERNEST (Exploration Rover for Navigating Extreme Terrain)~\cite{endurance2025missionProgress}, illustrated in \cref{fig:ernest}. It fuses visual and inertial sensors for state estimation. Hence, it is important to study the characteristics of \ac{vio} with a focus on low frame rates, thus reducing the need for frequent image capturing and onboard LED strobing, minimizing computational and energy requirements.

\par We propose a scene-adaptive, \ac{bev}-perspective, feature-matching framework to increase inlier feature matches across a larger baseline between subsequent camera images. Our proposed method integrates into the feature-matching frontend module of the xVIO~\cite{delaune2020xvio}. We conduct detailed experiments in simulation and in the field using a half-scale rover.

\par We find that by employing \ac{bev}, reliable \ac{vio} can be achieved at frame rates as low as \SI{0.25}{Hz}. In spatial terms, this corresponds to $\approx$\SI{1}{m} distance traveled between two consecutive images. To emphasize the challenge in the problem we tackle, it is important to understand that in missions like the Mars 2020 Perseverance, where the operation is limited to daytime, the image updates were performed every \SI{1}{m} due to computational limitations. For fair comparison, the scale-compensated baseline between images for the Perseverance rover is approximately \SI{0.5}{m}, accounting for the half-scale prototype ERNEST, with the camera mounted at roughly half the height of the Perseverance and Endurance rovers. Our findings indicate a promising two-fold improvement in the baseline between images over the scale-compensated baseline for the Perseverance rover, while operating across day and night-time, with up to six times higher average speeds.

\par Further, to compare against earth-based flying systems with \ac{vio}, consider an aerial robot with a camera operating at \SI{20}{Hz} and flying at \SI{20}{m/s}, resulting in an image-to-image baseline of \SI{1}{m}. In this case, the flying system benefits from \ac{imu} biases evolving unconstrained for only \SI{0.05}{s}. By contrast, in the problem we address with rover traversing at average speeds of \SI{0.25}{m/s}, the \ac{imu} biases evolve unconstrained for approximately \SI{4.0}{s}, which is significantly longer.

\par The remainder of this paper is organized as follows: Section II reviews related work, Section III presents our scene-adaptive \ac{bev} method, Section IV describes VIO integration, and Section V describes the simulation and field experiments, while Section VI draws conclusions.

%% file: tex/2_RelatedWork.tex
Vision-based state estimation, and \ac{vio} in particular, have evolved significantly over the past two decades. Seminal contributions to \ac{vio} can be broadly categorized into loosely-coupled methods (fusing independent visual and inertial estimates)~\cite{weiss2012real}, tightly-coupled methods (joint optimization of both modalities)~\cite{leutenegger2015keyframe}, filter-based frameworks (e.g., Extended Kalman Filters)~\cite{mourikis2007multi}, and optimization-based systems (e.g., sliding-window bundle adjustment)~\cite{qin2018vins}. On earth-based systems, often the choice of \ac{vio} can be across this broad spectrum, while for space applications, the early solutions have been loosely-coupled \ac{vo} with \ac{imu} or wheel odometry, while recent solutions are also based on bundle-adjustment, and tightly-coupled filter-based methods.
\par For earth-based systems, the Multi-State Constraint Kalman Filter (MSCKF)~\cite{mourikis2007multi} is a foundational filter-based method that encodes feature constraints in the null space of the feature Jacobian, decoupling updates from feature count and enabling accurate, efficient VIO. Building on this, OpenVINS~\cite{openvins} further refines the MSCKF framework. In the optimization family, OKVIS~\cite{leutenegger2015keyframe} introduced keyframe-based nonlinear optimization with marginalization, improving robustness. Meanwhile, ROVIO~\cite{bloesch2015robust} employs an iterated EKF with a semi-direct, patch-based photometric error and a robocentric formulation, improving initialization robustness and performance in textureless environments.

These methods have been extensively tested on Earth under various conditions, including dynamic motions and varying illumination.
\ac{vio} is applied across a wide range of domains, including robotics, where it has been deployed on aerial, ground, and underwater robots \cite{Cadena16tro-SLAMfuture}. Naturally, for space applications, the vision-based state-estimation method shares the same fundamental components, while mission-specific requirements like computational efficiency and energy constraints further tailor the solutions.

Among one of the early works, the Mars Exploration Rovers (MER) \cite{maimone2007two} use state estimation from wheel odometry and frame-to-frame feature tracking based \ac{vo}. They detect Harris corners and perform stereo matching to obtain the three-dimensional location and corresponding covariances. The rover then moves, and the past feature locations are re-projected into the image frame. A correlation-based search is used to match the features. The estimation is performed in two steps: first, with a least-square fit with RANSAC, followed by Maximum likelihood estimation. Towards lunar applications, some works proposed a downward-facing stereo camera with self-illumination~\cite{wagner2012visual} where they use a downwards facing stereo camera pair with self-illumination, tracking features to perform state-estimation. While other works proposed an image-enhancement method over a stereo visual odometry with Harris corner detection and feature tracking for motion estimation~\cite{li2013visual}.

In more recent developments, the Mars Helicopter~\cite{bayard2019vision} used an \ac{ekf}-based \ac{vio}. The full sensor stack includes a downwards-facing camera, a single-beam range-finder, and an \ac{imu}. The method uses FAST~\cite{rosten2006machine} features and KLT~\cite{tomasi1991detection} tracking with gyro-derotation, followed by RANSAC outlier rejection. The Yutu-2 lunar rover state estimation is based on SURF~\cite{bay2006surf} features matching and manually selected features for stereo bundle adjustment based on \ac{imu}-assisted \ac{vo}. A recent work, LuVo~\cite{luvo2025} utilized a \ac{bev} perspective for enhancing feature matching across large translations between images. Their solution proposes using learning-based feature matching with a remote, earth-based server for \ac{gpu} computations, whereas the proposed method runs onboard the rover. Further, they obtain the \ac{bev} perspective by assuming flat terrain and using rover attitude to compute the warping.

Most of the prior methods targeting space deployment have been developed for slow-moving rovers (e.g. \SI{0.04}{m/s} for the Mars 2020 Perseverance rover) or flying robots, which can be fast but function with high overlap between subsequent images due to higher altitude from the ground. To this end, we tackle the problem of a fast-moving planetary rover $(\approx \SI{0.25}{cm/s})$ on challenging terrain for long-term traversal across day and night with an onboard real-time solution. Our approach uses real-time three-dimensional terrain information from dense stereo to identify the plane normal for the BEV projection. While feature matching in \ac{bev} enables us to reduce the image \ac{fps}, our \ac{vio} integrated solution outputs high-rate state estimates, necessary for fast lunar rover navigation.

%% file: tex/3_BEV.tex
In this section, we describe the insight behind \ac{bev} and how we obtain the scene-adaptive \ac{bev} perspective, given the image from the rover's navigation camera and the stereo depth point cloud. We also discuss feature detection and matching after obtaining the images in the \ac{bev} perspective. The derivations in this section stem from the two-dimensional projective geometry described in \cite{hartley2003multiple}. 
\par The key insight is that in \ac{bev} perspective, the scene appearance remains consistent between frames; features primarily translate laterally in the image plane rather than undergoing large appearance changes. This greatly reduces descriptor mismatch compared to the \ac{is}, where inter-frame appearance change is significantly larger.

\begin{figure}
    \centering
    \input{Tikz/BEV_diagram}
    \caption{\small Illustration of the perspectives of the Image Space and Bird's Eye View, alongside the coordinate frames used in the proposed formulation.}
    \label{fig:bev_diagram}
    \vspace{-10px}
\end{figure}
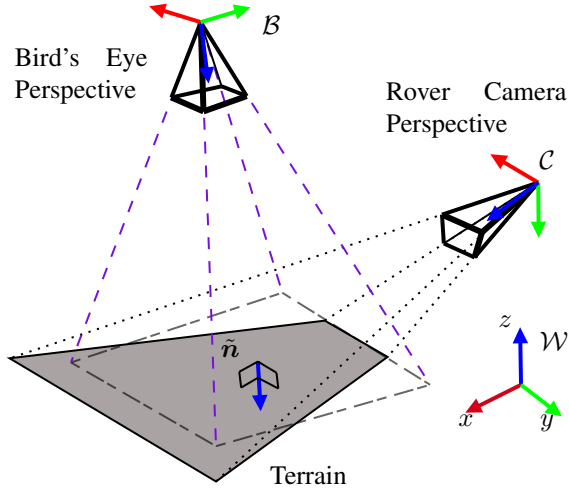

\subsection{Notations}
\label{subsec:notations}
\begin{itemize}
    \item $\mathcal{C}$: Rover left stereo navigation camera coordinate frame, with the $z$-axis along optical axis.
    \item $\mathcal{I}$: Rover \ac{imu} coordinate frame.
    \item $\mathcal{B}$: Virtual \ac{bev} camera coordinate frame, with the $z$-axis perpendicular to the ground plane.
    \item $\mathcal{W}$: World coordinate frame, with the $z$-axis parallel to the gravity vector and pointing upwards.
    \item $(H, W)$: Height and width of the camera image.
    \item $(H_{\text{BEV}}, W_{\text{BEV}})$: Height and width of the \ac{bev} image.
    \item Vectors are denoted by boldface symbols.
\end{itemize}

\subsection{Bird's-Eye-View transformation}
In our formulation, the rover navigation camera observes the scene (\cref{fig:bev_diagram}). We assume that the scene is approximately flat and a plane (parametrized by the normal unit vector $\tilde{\boldsymbol{n}}$) can be fit to this terrain. We compute a homography transformation $M$, defined as:
\begin{equation}
    M = K R_{\mathcal{B},\mathcal{C}} K^{-1},
\end{equation}
where $K$ is the camera intrinsic matrix and $R_{\mathcal{B},\mathcal{C}} \in SO(3)$ denotes the rotation matrix from $\mathcal{C}$ to $\mathcal{B}$.  The intrinsics $K$ are obtained through camera calibration, and we assume distortion-free, rectified input images.
\par The rotation matrix $R_{\mathcal{B},\mathcal{C}}$ is derived using the plane normal $\tilde{\boldsymbol{n}}$ and the axes of $\mathcal{C}$. Let $\tilde{\boldsymbol{x}}_{\mathcal{C}}$ and $\tilde{\boldsymbol{z}}_{\mathcal{C}}$ denote the $x$- and $z$-axes of frame $\mathcal{C}$, then $R_{\mathcal{B},\mathcal{C}}$ can be given by:
\begin{equation}
    R_{\mathcal{B},\mathcal{C}} = \begin{bmatrix}
        \big( \tilde{\boldsymbol{n}} \times \tilde{\boldsymbol{z}}_{\mathcal{C}} \big) \times \tilde{\boldsymbol{x}}_{\mathcal{C}} &
        \tilde{\boldsymbol{n}} \times \tilde{\boldsymbol{z}}_{\mathcal{C}} &
        \tilde{\boldsymbol{n}}
    \end{bmatrix}.
\end{equation}
\par To control the resolution of the BEV image, we apply a scaling operation with matrix
\begin{equation}
    S = \begin{bmatrix}
        s & 0 & 0 \\
        0 & s & 0 \\
        0 & 0 & 1
    \end{bmatrix},
\end{equation}
where $s$ is the scaling factor. Furthermore, a $2D$ translation is applied in image coordinates using a translation matrix $T$ to align the midpoint of the lower edge of the input image $(H, W/2)$ with the bottom of the BEV frame $(H_{BEV}, W_{BEV}/2)$. Scaling is a tunable heuristic; it ensures fine resolution for nearby points. In our case, we set $s=0.8$. Translation minimizes unmapped regions manifesting as pixels with $0$ intensity. The complete homography matrix $M^{*}$ is then given by
\begin{equation}
    M^{*}= TSM.
\end{equation}

\subsection{Plane normal estimation}
\label{subsec:plane_nor_est}
We estimate the ground plane normal by fitting a plane to the 3D points obtained from the stereo depth image $D \in \mathbb{R}^{H \times W}$ using least squares with RANSAC. To reduce computation, we downsample the depth image to $(H/g, W/g)$ points, where $g$ is the downsampling factor. The selected points are the centers of non-overlapping patches of size $(g,g)$. Using the pixel coordinates, depth values, and $K$, we obtain the corresponding 3D points in $\mathcal{C}$, denoted as $\mathbf{P}_{\mathcal{C}}$, and estimate the plane normal $\tilde{\boldsymbol{n}}_{\mathcal{C}}$.
\par We transform $\tilde{\boldsymbol{n}}_{\mathcal{C}}$ into the world frame:
\begin{equation}
    \tilde{\boldsymbol{n}}_{\mathcal{W}} = R_{\mathcal{W},\mathcal{C}} \, \tilde{\boldsymbol{n}}_{\mathcal{C}},
\end{equation}
where $R_{\mathcal{W},\mathcal{C}}$ is obtained from the rover’s orientation estimate $\mathbf{q}_{\mathcal{W},\mathcal{C}}$ provided by the state-estimation filter (described in \cref{sec: vio}).
\begin{figure*}[ht!]
    \centering    \includegraphics[width=0.99\linewidth]{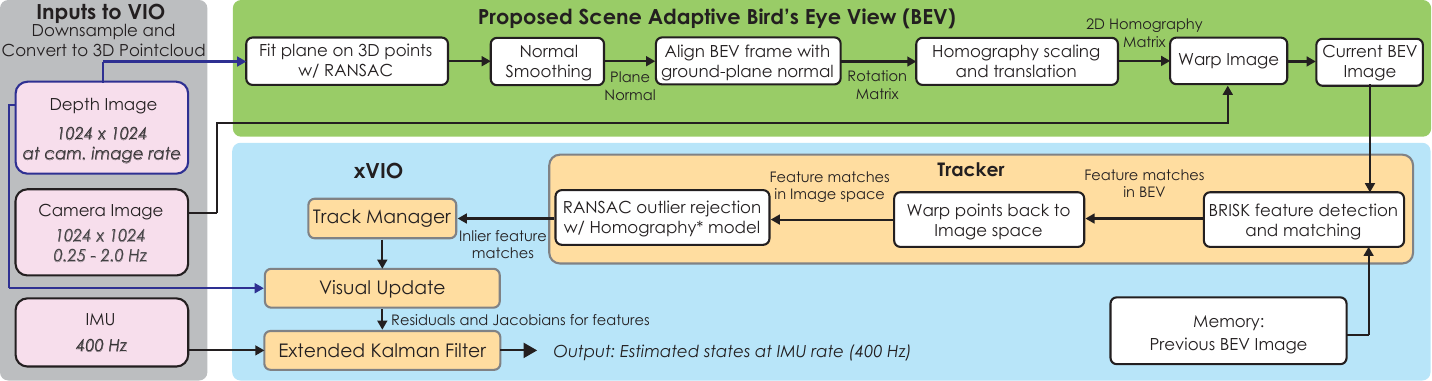}
    \caption{\small Overview for the proposed method and its integration into \ac{vio}. The blocks in white color illustrate the parts of the proposed approach, and blocks in yellow color illustrate, modules of the baseline method xVIO\cite{delaune2020xvio}.}
    \label{fig:bev_pipeline}
    \vspace{-10px}
\end{figure*}
To smoothen the computed normal in the world frame, we maintain a sliding window of normals $\mathbb{N} = \{ \tilde{\boldsymbol{n}}_{\mathcal{W},i} \mid t-m \leq i \leq t \}$ and compute their moving average $\tilde{\boldsymbol{n}}_{\mathcal{W}}^*$. The averaged normal is then transformed back into $\mathcal{C}$ as follows:
\begin{equation}
    \tilde{\boldsymbol{n}}_{\mathcal{C}}^* = R_{\mathcal{W},\mathcal{C}}^{T} \tilde{\boldsymbol{n}}^*_{\mathcal{W}}.
\end{equation}

\subsection{Feature matching in BEV}
Given input images, $I_{t-1}$ and $I_{t}$ (each of dimensions $H\times W$) from the navigation camera at times $t-1$ and $t$ respectively. The \ac{bev} images are obtained by warping the camera images using the homography at the corresponding times:
\begin{subequations}
\begin{align}
I_{t,\text{BEV}} & = \phi\left(I_t, M^*_{t}\right), \\
I_{t-1,\text{BEV}} & = \phi\left(I_{t-1}, M^*_{t-1}\right),
\end{align}
\end{subequations}
where $\phi$ is the warping function defined from a pixel coordinate $\boldsymbol{p}_{\text{IS}}$ in \ac{is} image to pixel coordinate in  $\boldsymbol{p}_{\text{BEV}}$ in \ac{bev} image and is given by
\begin{equation}
    \boldsymbol{p}_{\text{BEV}}=M\ \boldsymbol{p}_{\text{IS}}
\end{equation}
for a given homography matrix $M$.
The unmapped regions are assigned zero intensity. The warped \ac{bev} images are of dimensions of $\left(H_{\text{BEV}}\times W_{\text{BEV}}\right)$. To avoid spurious detections, feature extraction is masked to exclude boundaries between valid and unmapped areas.
\par In our formulation, we employ BRISK~\cite{brisk} features and descriptors. This choice is motivated by its robustness to illumination variations across images captured from large baselines, especially compared to optical-flow based feature tracking using KLT\cite{tomasi1991detection} (used in xVIO~\cite{delaune2020xvio}), which assumes constant intensity. KLT is suitable for high \ac{fps} image streams. Additionally, BRISK was chosen due to the computational efficiency of its binary descriptors over floating-point descriptors~\cite{lowe2004distinctive,bay2006surf}. We detect BRISK features on $I_{t,\text{BEV}}$ and $I_{t-1,\text{BEV}}$ and match them with normalized Hamming distance as described in~\cite{brisk}.

\par The matched features contain outliers, and to reject these, we use a homography model with RANSAC. During this step, the estimated homography is also decomposed to recover the supporting plane normal, which is compared against the ideal flat-terrain normal for consistency verification. This yields a robust set of matched visual features. Although the feature detection and matching operations are performed in the \ac{bev} space, we transform the points back to \ac{is} using the inverse of the homography matrices corresponding to each timestep. This enables us to detect and match features in BEV, without changing the downstream formulation of \ac{vio}. \cref{fig:bev_pipeline} showcases the overall framework.

%% file: Tikz/BEV_diagram.tex
\tikzset{every picture/.style={line width=0.75pt}} 

\begin{tikzpicture}[x=0.75pt,y=0.75pt,yscale=-0.70,xscale=0.70]
\draw  [fill={rgb, 255:red, 169; green, 163; blue, 163 }  ,fill opacity=1 ] (335,227) -- (380,253) -- (257,342.33) -- (109,254) -- cycle ;
\draw [color={rgb, 255:red, 0; green, 0; blue, 255 }  ,draw opacity=1 ][line width=1.5]    (287,256.33) -- (288.78,288.01) ;
\draw [shift={(289,292)}, rotate = 266.79] [fill={rgb, 255:red, 0; green, 0; blue, 255 }  ,fill opacity=1 ][line width=0.08]  [draw opacity=0] (11.61,-5.58) -- (0,0) -- (11.61,5.58) -- cycle    ;
\draw  [dash pattern={on 0.84pt off 2.51pt}]  (335,227) -- (488,128) ;
\draw  [dash pattern={on 0.84pt off 2.51pt}]  (488,128) -- (380,253) ;
\draw  [dash pattern={on 0.84pt off 2.51pt}]  (257,342.33) -- (488,128) ;
\draw  [dash pattern={on 0.84pt off 2.51pt}]  (488,128) -- (109,254) ;
\draw [color={rgb, 255:red, 0; green, 0; blue, 0 }  ,draw opacity=0.6 ][fill={rgb, 255:red, 226; green, 27; blue, 27 }  ,fill opacity=1 ] [dash pattern={on 3.75pt off 3pt on 7.5pt off 1.5pt}]  (153,257) -- (173.99,269.11) -- (257,317) ;
\draw [color={rgb, 255:red, 0; green, 0; blue, 0 }  ,draw opacity=0.6 ][fill={rgb, 255:red, 226; green, 27; blue, 27 }  ,fill opacity=1 ] [dash pattern={on 3.75pt off 3pt on 7.5pt off 1.5pt}]  (153,257) -- (234.47,230.11) -- (304.5,207) ;
\draw [color={rgb, 255:red, 0; green, 0; blue, 0 }  ,draw opacity=0.6 ][fill={rgb, 255:red, 226; green, 27; blue, 27 }  ,fill opacity=1 ] [dash pattern={on 3.75pt off 3pt on 7.5pt off 1.5pt}]  (257,317) -- (410.5,273) ;
\draw [color={rgb, 255:red, 0; green, 0; blue, 0 }  ,draw opacity=0.6 ][fill={rgb, 255:red, 226; green, 27; blue, 27 }  ,fill opacity=1 ] [dash pattern={on 3.75pt off 3pt on 7.5pt off 1.5pt}]  (304.5,207) -- (405.19,269.69) -- (410.5,273) ;

\draw [color={rgb, 255:red, 121; green, 16; blue, 213 }  ,draw opacity=1 ] [dash pattern={on 4.5pt off 4.5pt}]  (246,13) -- (257,317) ;
\draw [color={rgb, 255:red, 121; green, 16; blue, 213 }  ,draw opacity=1 ] [dash pattern={on 4.5pt off 4.5pt}]  (246,13) -- (153,257) ;
\draw [color={rgb, 255:red, 121; green, 16; blue, 213 }  ,draw opacity=1 ][fill={rgb, 255:red, 144; green, 19; blue, 254 }  ,fill opacity=1 ] [dash pattern={on 4.5pt off 4.5pt}]  (246,13) -- (410.5,273) ;
\draw [color={rgb, 255:red, 121; green, 16; blue, 213 }  ,draw opacity=1 ] [dash pattern={on 4.5pt off 4.5pt}]  (246,13) -- (304.5,207) ;
\draw [line width=1.5]    (246,13) -- (225,67) ;
\draw [line width=2.25]    (248,77) -- (225,67) ;
\draw [line width=2.25]    (279,66) -- (248,77) ;
\draw [line width=1.5]    (261,59) -- (279,66) ;
\draw [line width=1.5]    (225,67) -- (261,59) ;
\draw [line width=2.25]    (246,13) -- (248,77) ;
\draw [line width=1.5]    (246,13) -- (279,66) ;
\draw    (246,13) -- (261,59) ;
\draw [line width=1.5]    (419,151) -- (488,128) ;
\draw [line width=1.5]    (419,151) -- (420.33,171.67) ;
\draw [line width=2.25]    (419,151) -- (447.67,163.67) ;
\draw [line width=1.5]    (420.33,171.67) -- (442.33,181.67) ;
\draw [line width=2.25]    (442.33,181.67) -- (447.67,163.67) ;
\draw [line width=2.25]    (488,128) -- (447.67,163.67) ;
\draw [line width=1.5]    (488,128) -- (442.33,181.67) ;
\draw    (488,128) -- (420.33,171.67) ;
\draw [color={rgb, 255:red, 0; green, 0; blue, 255 }  ,draw opacity=1 ][line width=1.5]    (246,13) -- (251.16,53.03) ;
\draw [shift={(251.67,57)}, rotate = 262.66] [fill={rgb, 255:red, 0; green, 0; blue, 255 }  ,fill opacity=1 ][line width=0.08]  [draw opacity=0] (11.61,-5.58) -- (0,0) -- (11.61,5.58) -- cycle    ;
\draw [color={rgb, 255:red, 255; green, 0; blue, 0 }  ,draw opacity=1 ][line width=1.5]    (246,13) -- (212.82,2.84) ;
\draw [shift={(209,1.67)}, rotate = 17.03] [fill={rgb, 255:red, 255; green, 0; blue, 0 }  ,fill opacity=1 ][line width=0.08]  [draw opacity=0] (11.61,-5.58) -- (0,0) -- (11.61,5.58) -- cycle    ;
\draw [color={rgb, 255:red, 0; green, 255; blue, 0 }  ,draw opacity=1 ][line width=1.5]    (246,13) -- (277.83,3.48) ;
\draw [shift={(281.67,2.33)}, rotate = 163.35] [fill={rgb, 255:red, 0; green, 255; blue, 0 }  ,fill opacity=1 ][line width=0.08]  [draw opacity=0] (11.61,-5.58) -- (0,0) -- (11.61,5.58) -- cycle    ;
\draw [color={rgb, 255:red, 0; green, 255; blue, 0 }  ,draw opacity=1 ][line width=1.5]    (488,128) -- (488.3,163.67) ;
\draw [shift={(488.33,167.67)}, rotate = 269.52] [fill={rgb, 255:red, 0; green, 255; blue, 0 }  ,fill opacity=1 ][line width=0.08]  [draw opacity=0] (11.61,-5.58) -- (0,0) -- (11.61,5.58) -- cycle    ;
\draw [color={rgb, 255:red, 255; green, 0; blue, 0 }  ,draw opacity=1 ][line width=1.5]    (488,128) -- (457.11,109.7) ;
\draw [shift={(453.67,107.67)}, rotate = 30.64] [fill={rgb, 255:red, 255; green, 0; blue, 0 }  ,fill opacity=1 ][line width=0.08]  [draw opacity=0] (11.61,-5.58) -- (0,0) -- (11.61,5.58) -- cycle    ;
\draw [color={rgb, 255:red, 0; green, 0; blue, 255 }  ,draw opacity=1 ][line width=1.5]    (488,128) -- (454.2,153.27) ;
\draw [shift={(451,155.67)}, rotate = 323.21] [fill={rgb, 255:red, 0; green, 0; blue, 255 }  ,fill opacity=1 ][line width=0.08]  [draw opacity=0] (11.61,-5.58) -- (0,0) -- (11.61,5.58) -- cycle    ;
\draw [color={rgb, 255:red, 0; green, 0; blue, 255 }  ,draw opacity=1 ][line width=1.5]    (475.67,273) -- (475.06,235.67) ;
\draw [shift={(475,231.67)}, rotate = 89.08] [fill={rgb, 255:red, 0; green, 0; blue, 255 }  ,fill opacity=1 ][line width=0.08]  [draw opacity=0] (11.61,-5.58) -- (0,0) -- (11.61,5.58) -- cycle    ;
\draw [color={rgb, 255:red, 208; green, 2; blue, 27 }  ,draw opacity=1 ][fill={rgb, 255:red, 228; green, 18; blue, 44 }  ,fill opacity=1 ][line width=1.5]    (475.67,273) -- (439.24,291.21) ;
\draw [shift={(435.67,293)}, rotate = 333.43] [fill={rgb, 255:red, 208; green, 2; blue, 27 }  ,fill opacity=1 ][line width=0.08]  [draw opacity=0] (11.61,-5.58) -- (0,0) -- (11.61,5.58) -- cycle    ;
\draw [color={rgb, 255:red, 0; green, 255; blue, 0 }  ,draw opacity=1 ][line width=1.5]    (475.67,273) -- (501.83,293.22) ;
\draw [shift={(505,295.67)}, rotate = 217.69] [fill={rgb, 255:red, 0; green, 255; blue, 0 }  ,fill opacity=1 ][line width=0.08]  [draw opacity=0] (11.61,-5.58) -- (0,0) -- (11.61,5.58) -- cycle    ;
\draw    (301.67,275.67) -- (287,268) ;
\draw    (273.67,263.67) -- (287,256.33) ;
\draw    (273.67,263.67) -- (275,275) ;
\draw    (300.33,264.33) -- (301.67,275.67) ;
\draw    (287,256.33) -- (300.33,264.33) ;
\draw    (287,268) -- (275,275) ;

\draw (288,6.67) node [anchor=north west][inner sep=0.75pt]   [align=left] {$\displaystyle \mathcal{B}$};
\draw (486.67,103.33) node [anchor=north west][inner sep=0.75pt]   [align=left] {$\displaystyle \mathcal{C}$};
\draw (428.67,292.33) node [anchor=north west][inner sep=0.75pt]   [align=left] {$\displaystyle x$};
\draw (487.33,292.33) node [anchor=north west][inner sep=0.75pt]   [align=left] {$\displaystyle y$};
\draw (457.33,221) node [anchor=north west][inner sep=0.75pt]   [align=left] {$\displaystyle z$};
\draw (160.33,51.67) node   [align=left] {\begin{minipage}[lt]{50pt}\setlength\topsep{0pt}
{\fontfamily{ptm}\selectfont Bird's Eye Perspective}
\end{minipage}};
\draw (443.67,75.67) node   [align=left] {\begin{minipage}[lt]{68pt}\setlength\topsep{0pt}
{\fontfamily{ptm}\selectfont Rover Camera Perspective}
\end{minipage}};
\draw (360.67,337.67) node   [align=left] {\begin{minipage}[lt]{68pt}\setlength\topsep{0pt}
{\fontfamily{ptm}\selectfont Terrain}
\end{minipage}};
\draw (271.33,246.67) node   [align=left] {\begin{minipage}[lt]{11.33pt}\setlength\topsep{0pt}
$\displaystyle \tilde{\boldsymbol{n}}$
\end{minipage}};
\draw (487,233) node [anchor=north west][inner sep=0.75pt]   [align=left] {$\displaystyle \mathcal{W}$};

\end{tikzpicture}

%% file: tex/4_VIOIntegration.tex
We use xVIO~\cite{delaune2020xvio} as the underlying \ac{vio} framework. 
The input sensor modalities include the camera images, the \ac{imu} measurements, and the range measurements. In our case, the image is from the left camera of the stereo pair, and the range measurements come from the stereo matching depth image. The \ac{imu} measurements are from a MEMS \ac{imu} mounted rigidly with respect to the cameras. We use the same notations for coordinate frame from \cref{subsec:notations}. The state vector in xVIO is defined as  
\begin{equation}
    \mathbf{s} = 
    \begin{bmatrix}
        \mathbf{s}_{I}^{T} & \mathbf{s}_{V}^{T}
    \end{bmatrix}^{T},
\end{equation}
where $\mathbf{s}_{I}$ denotes the inertial state associated with the IMU, and $\mathbf{s}_{V}$ represents the visual state corresponding to tracked features.  

The inertial state $\mathbf{s}_{I} \in \mathbb{R}^{16}$ comprises the position, velocity, orientation, gyroscope biases, and accelerometer biases of the IMU, and can be written as  
\begin{equation}
    \mathbf{s}_{I} =
    \begin{bmatrix}
        \mathbf{p}^{T} & \mathbf{v}^{T} & \mathbf{q}^{T} & \mathbf{b}_g^{T} & \mathbf{b}_a^{T}
    \end{bmatrix}^{T}.
\end{equation}

The error-state is partitioned in the same manner (inertial and visual components), which enables EKF linearization and covariance propagation.  

Internally, xVIO is structured into three main modules: \textit{tracker}, \textit{track manager}, and \textit{visual update}.  

\begin{itemize}
    \item \textbf{Tracker:} Implements frame-to-frame feature tracking, serving as the visual front-end. Our proposed \ac{bev}-based feature matching is integrated within this module.  
    \item \textbf{Track manager:} Maintain keyframes, and triggers their creation based on the number of tracked features output by the tracker.  
    \item \textbf{Visual update:} Handles computation of Jacobians and residuals for EKF update associated with features.
\end{itemize}
Given the orientation of the IMU in the world frame, $\mathbf{q}$, and the camera-to-IMU transformation, $\mathbf{q}_{\mathcal{C,I}}$, we can compute the orientation of the camera in the world frame, $\mathbf{q}_{\mathcal{W,C}}$, which is then used for plane normal estimation (see \cref{subsec:plane_nor_est}). Specifically, we use the predicted orientation, $\hat{\mathbf{q}}$, obtained from the filter prediction step, as it is required prior to the image measurement update. For further details on xVIO, the reader is referred to \cite{delaune2020xvio}.

%% file: tex/5_experiments.tex
This section presents a comprehensive evaluation of our proposed method against the baseline \ac{vio} approach, using both qualitative and quantitative analyses. Here, baseline \ac{vio} refers to a modified version of xVIO~\cite{delaune2020xvio} with a BRISK feature matching front-end instead of KLT~\cite{tomasi1991detection}, and a stereo depth image for range measurements instead of a single laser range-finder. First, we assess the performance gains achieved by employing the \ac{bev} perspective compared to the \ac{is} used in the baseline for feature matching. Second, we assess the possible reduction in image \ac{fps} (i.e., to enable feature matching at larger baselines) achieved by utilizing \ac{bev} in the feature matching front-end of \ac{vio}. Prior to the evaluation, we present the experimental setup, commencing with the rover platform employed in this study.

\subsection{ERNEST rover prototype}
Our studies were conducted using data collected from simulated and real-world environments, with a half-scale Endurance rover prototype named ``ERNEST" serving as the robot platform. ERNEST is a four-wheeled surface mobility rover with a suspension that can be configured to be actively controlled or passive. The active suspension allows the rover to navigate more extreme terrains and slopes up to \SI{30}{\degree}. For the Endurance mission prototype testing, the rover was configured to use its passive suspension given the anticipated terrain. In our experiments, the rover maintains an average velocity of \SI{0.24}{m/s}.

The state-estimation sensor suite includes a synchronized global shutter stereo camera pair and a VectorNav VN100 MEMS \ac{imu}, both mounted on a fixed mast at the front of the rover. The stereo camera pair is pitched down by \SI{30}{\degree}, and the \ac{fov} for each camera is \SI{90}{\degree}$\times$\SI{90}{\degree}. The rover's autonomy stack runs on the onboard computer, with high-level commands provided from a ground control station. The camera is triggered at \SI{2}{Hz}, while the IMU is sampled at \SI{400}{Hz}. To emulate a low frame rate image stream, we developed a pre-processing script that skips a specified number of images to achieve the desired image \ac{fps}. The cameras were calibrated using the Kalibr toolbox. The cameras outputs grayscale images of dimension $1024\times1024$ pixels. \cref{fig:ernest} shows the ERNEST rover in the field during the data collection in daytime and nighttime scenes. For nighttime perception, the rover uses a set of LED illuminators positioned below the stereo cameras on the mast. The LEDs irradiate red light for their highest optical response on camera pixels.

\subsection{Real-world experiment setup}
\begin{figure}[h!]
    \centering
    \begin{subfigure}[t]{0.15\textwidth} 
        \centering
        \includegraphics[width=\textwidth]{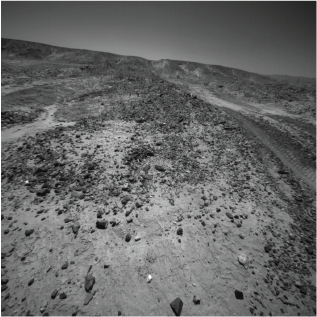}
        \caption{ Day}
        \label{fig:day_scene}
    \end{subfigure}
    \begin{subfigure}[t]{0.15\textwidth} 
        \centering
        \includegraphics[width=\textwidth]{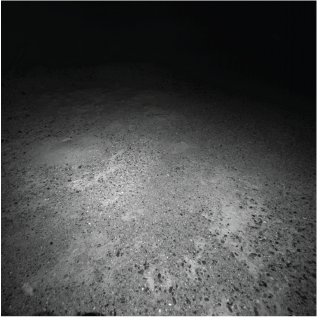}
        \caption{ Night}
        \label{fig:night_scene}
    \end{subfigure}
    \begin{subfigure}[t]{0.15\textwidth}
        \centering
        \includegraphics[width=\textwidth]{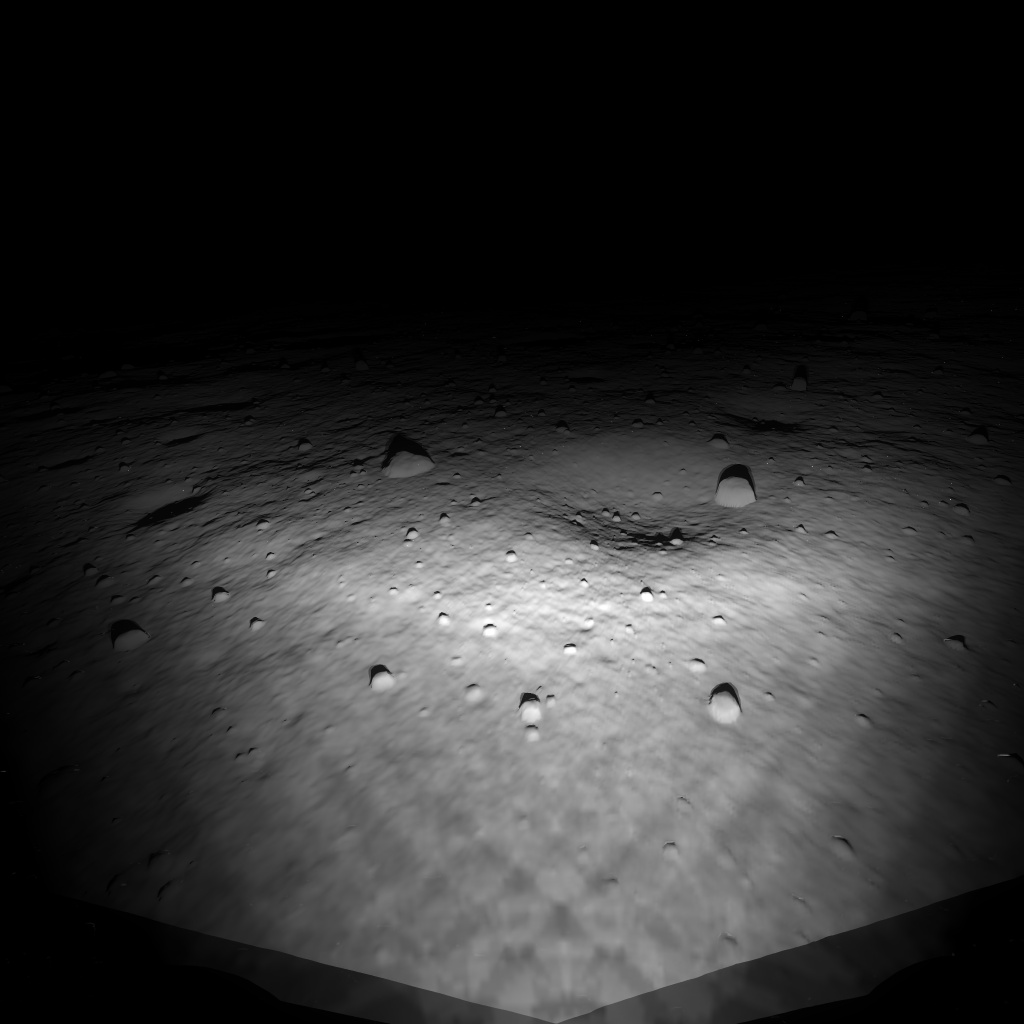}
        \caption{Night simulated}
        \label{fig:simulated_scene}
    \end{subfigure}
    \caption{\small Images from the rover left camera from the real-world day-time, real-world nighttime, and simulated nighttime environment.}
    \label{fig:rover_camera_view}
    \vspace{-10px}
\end{figure}

To analyze the state estimation, we collected extended data in the Plaster City desert region in California, USA. The scenes in the data qualitatively mimic a planetary environment. The navigation system continuously generates hazard-free trajectories for the rover to follow.  During the traversal, the rover covers paths of over hundreds of meters in both day and nighttime scenarios. Over its course, the rover traverses numerous regions, spanning sandy, rocky, and uneven terrain. \cref{tab:traj_lengths} shows the lengths of the different trajectories used in this evaluation. \cref{fig:rover_camera_view} shows the images from the rover left camera from the real-world day and nighttime, and for the nighttime simulated environment. The trajectories corresponding to real-world data from nighttime are \textit{Night-1} and \textit{Night-2}, while for daytime scenes are \textit{Day-1} and \textit{Day-2}.

\subsection{Lunar simulation setup}
To conduct our simulation analysis, we used the DARTS\cite{11068690} lunar simulator, which provides high-fidelity physics and image simulations for multiple lunar scenarios. This environment models both the complex terra-mechanics of traversing rocky terrains as well as the illumination effects from the sun and the rover's lights. The simulation uses an identical model of the Ernest rover, and the software stack is configured to function identically in both the simulator and real-world scenarios. The trajectory corresponding to simulated data is \textit{Night-3}.

\begin{table}[t]
\centering
\caption{Trajectory lengths (m) for different runs.}
\label{tab:traj_lengths}
\begin{tabular}{ccccc}
\toprule
Night-1 & Night-2 & Night-3 & Day-1 & Day-2 \\
\midrule
\SI{115}{m} & \SI{80}{m} & \SI{100}{m} & \SI{130}{m} & \SI{115}{m} \\
\bottomrule
\end{tabular}
\vspace{-10px}
\end{table}

\subsection{Effectiveness of Bird's Eye View}
Prior to evaluating the \ac{vio} performance, it is crucial to analyze the impact of reduced image frame-rate on feature matching using controlled study, independent from \ac{vio}. To this end, we evaluate the number of detected features, the ratio of inliers to total matched features, and the histogram of feature locations across varying frame rates.

Given a sequential image dataset, we select an image at time $t$ 
and match its features with another image captured at time $t+0.5k$ (factor of $0.5$ is used, given the recorded images are at $2$ 
\ac{fps}). 
By increasing the value of $k$, we effectively simulate a decreasing 
image frame rate. To obtain a statistical evaluation, $t \in \{t_1, t_2, 
\dots, t_n\}$ with $t_1 < t_2 < \cdots < t_n$.

\begin{figure}[h!]
    \centering
    \begin{subfigure}[t]{0.239\textwidth} 
        \centering
        \includegraphics[width=\textwidth]{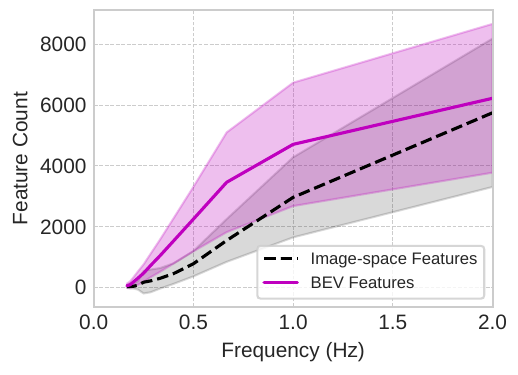}
        \caption{ Nighttime real-world.}
        \label{fig:nighttime_realworld}
    \end{subfigure}
    \hfill 
    \begin{subfigure}[t]{0.239\textwidth}
        \centering
        \includegraphics[width=\textwidth]{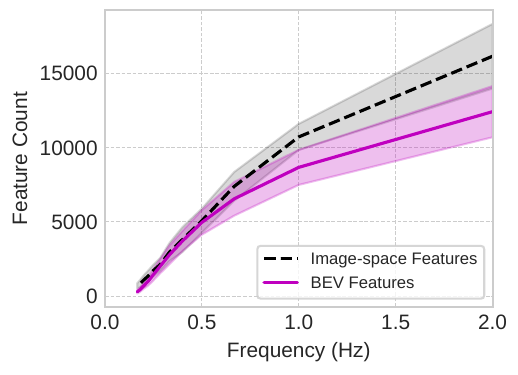}
        \caption{Daytime real-world.}
        \label{fig:daytime_realworld}
    \end{subfigure}
    \caption{\small Comparison of the number of features matched in \ac{is} vs. \ac{bev} in nighttime and daytime scenarios in real-world environment. The error bounds indicates maximum and minimum values.}
    \label{fig:ft_count_vs_freq}
    \vspace{-10px}
\end{figure}

\begin{figure}[h!]
    \centering
    \begin{subfigure}[t]{0.23\textwidth} 
        \centering
        \includegraphics[width=\textwidth]{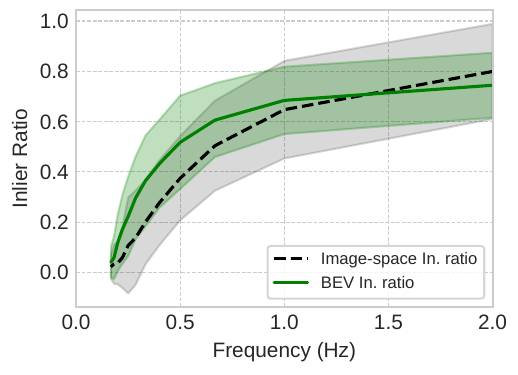}
        \caption{Nighttime real-world.}
        \label{fig:nighttime_realworld_ir}
    \end{subfigure}
    \hfill 
    \begin{subfigure}[t]{0.23\textwidth}
        \centering
        \includegraphics[width=\textwidth]{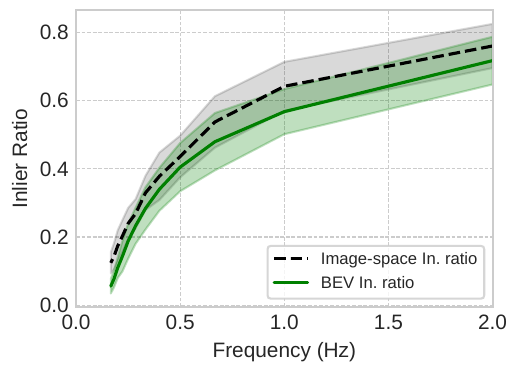}
        \caption{Daytime real-world.}
        \label{fig:daytime_realworld_ir}
    \end{subfigure}
    \caption{\small Comparison of the inlier ratio in \ac{is} vs. \ac{bev} in nighttime and daytime scenarios in real-world environment. The error bounds indicate maximum and minimum values.}
    \label{fig:inlier_ratio}
    \vspace{-10px}
\end{figure}

\cref{fig:ft_count_vs_freq} illustrates how the feature count varies with the image \ac{fps}. Notably, as the frequency decreases, a greater number of features are successfully matched using \ac{bev} perspective compared to \ac{is} at nighttime. We observe a similar trend in the inlier ratio in \cref{fig:inlier_ratio}, where the ratio is consistently higher for the \ac{bev} approach at lower frame rates than for the \ac{is} for nighttime. We also observe that for daytime scenes, both ``number of feature matched" and ``inlier ratio" are better for baseline; however, these matched features are distributed in the top part of the image, i.e., the far-off region of the scene. Due to the large distance of feature points from the rover, these matches are less informative for the rover's ego-motion, as they have lower parallax compared to nearby features, and they exhibit higher stereo depth estimation noise. The latter rationale is further elaborated in the next sub-section.

\subsection{Distribution of the matched features}
In our specific scenario, where the majority of the scene consists of a two-dimensional plane, a distribution in the lower region of the image indicates that the matched features are in the immediate vicinity of the rover. Conversely, a distribution in the upper region of the image denotes far-off features. Given this insight, we first showcase the effect of the matched features in \ac{bev} and \ac{is} in day and nighttime scenes at \SI{2.0}{Hz} and \SI{0.25}{Hz}. 

\begin{figure}[h!]
  \centering
  \begin{subfigure}{0.23\textwidth}
    \includegraphics[width=\linewidth]{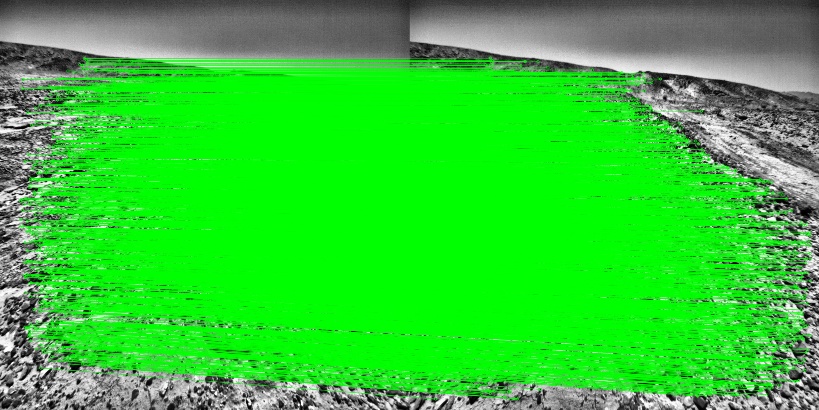}
    \caption{IS: \SI{2.0}{Hz}}
  \end{subfigure}
  \begin{subfigure}{0.23\textwidth}
    \includegraphics[width=\linewidth]{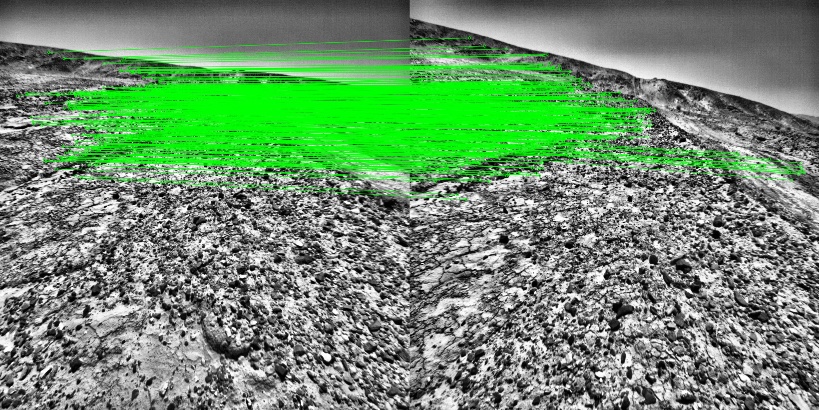}
    \caption{IS: \SI{0.25}{Hz}}
  \end{subfigure}
  \par\medskip 
  \begin{subfigure}{0.23\textwidth}
    \includegraphics[width=\linewidth]{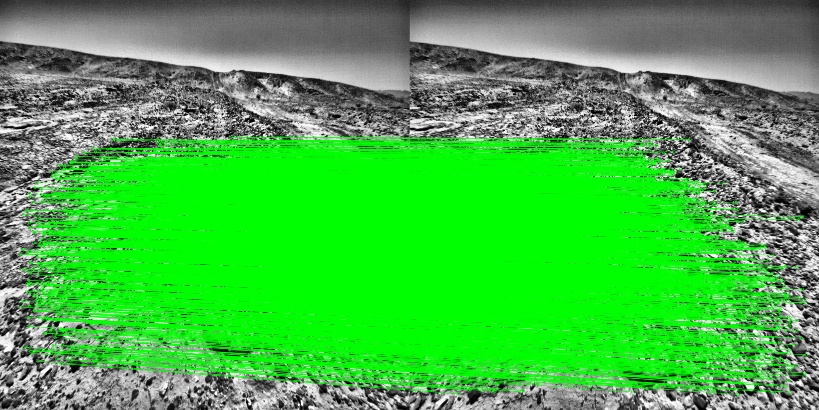}
    \caption{BEV: \SI{2.0}{Hz}}
  \end{subfigure}  
  \begin{subfigure}{0.23\textwidth}
    \includegraphics[width=\linewidth]{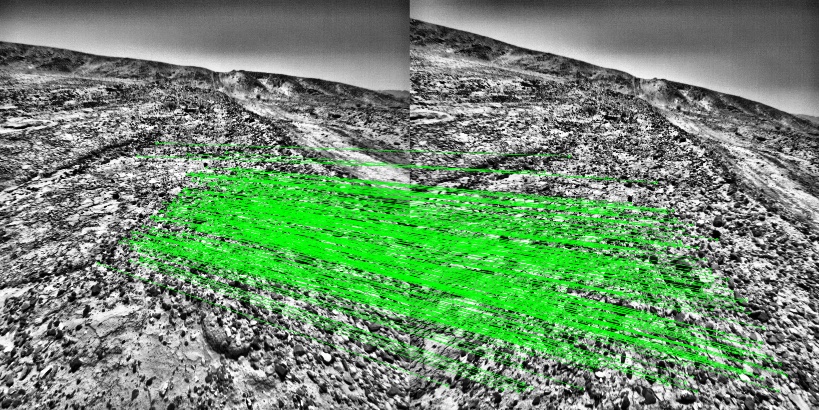}
    \caption{BEV: \SI{0.25}{Hz}}
  \end{subfigure}
   \par\medskip 
  \begin{subfigure}{0.23\textwidth}
    \includegraphics[width=\linewidth]{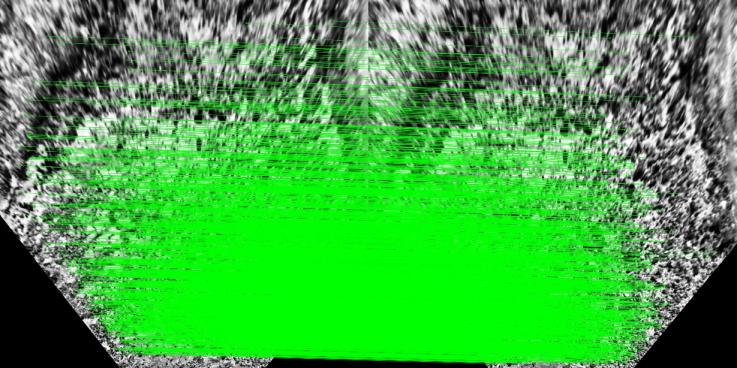}
    \caption{BEV$^{*}$: \SI{2.0}{Hz}}
  \end{subfigure}
  \begin{subfigure}{0.23\textwidth}
    \includegraphics[width=\linewidth]{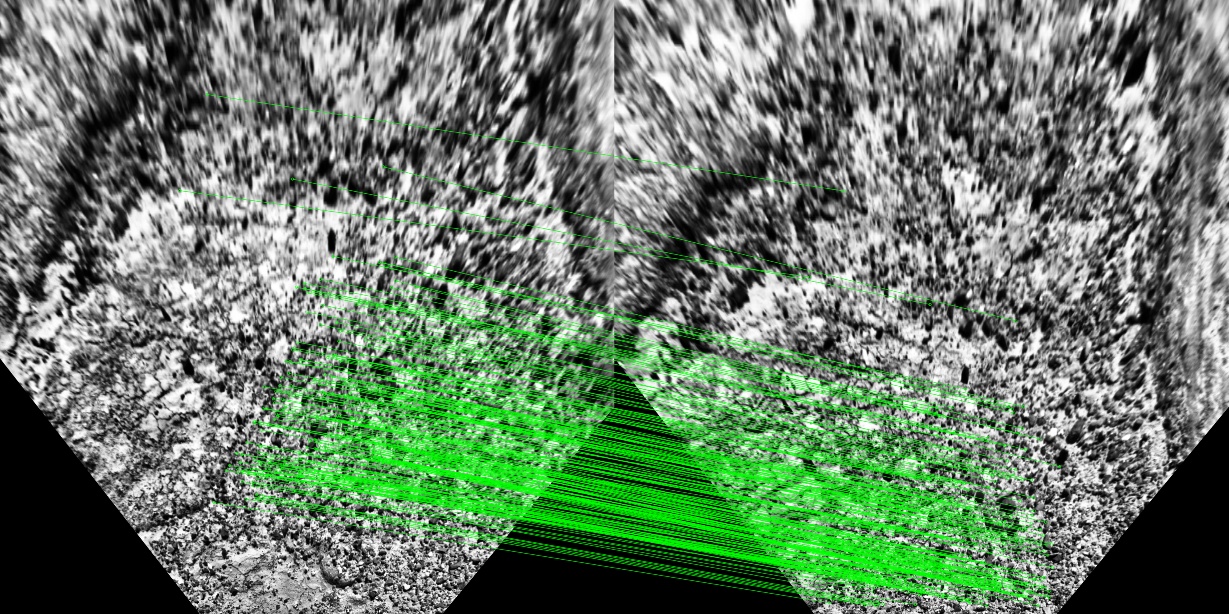}
    \caption{BEV$^{*}$: \SI{0.25}{Hz}}
  \end{subfigure}  
  \caption{\small Day-time feature matching comparison between feature matching in IS and BEV between images at \SI{2.0}{Hz} and \SI{0.25}{Hz}. Here, BEV denotes feature matched in BEV perspective and visualized in IS for comparability, and BEV$^{*}$ denotes that the features are matched in BEV and visualized in  BEV perspective.}
  \label{fig:day_ft_match}
  \vspace{-10px}
\end{figure}

\begin{figure}[h!]
  \centering
  \begin{subfigure}{0.23\textwidth}
    \includegraphics[width=\linewidth]{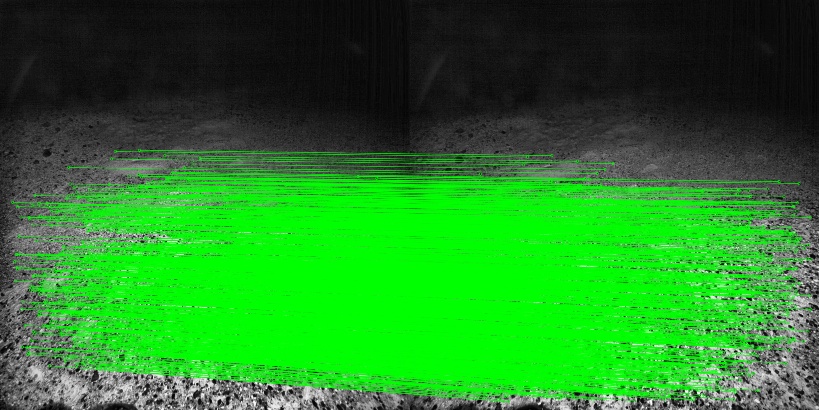}
    \caption{IS: \SI{2.0}{Hz}}
  \end{subfigure}
  \begin{subfigure}{0.23\textwidth}
    \includegraphics[width=\linewidth]{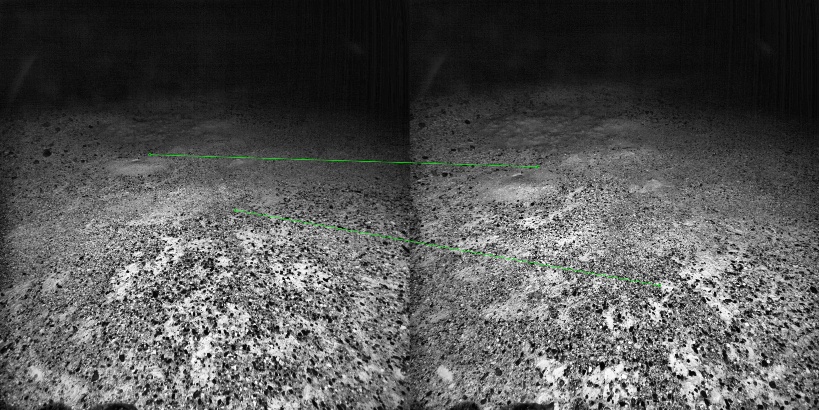}
    \caption{IS: \SI{0.25}{Hz}}
  \end{subfigure}
  \par\medskip 
  \begin{subfigure}{0.23\textwidth}
    \includegraphics[width=\linewidth]{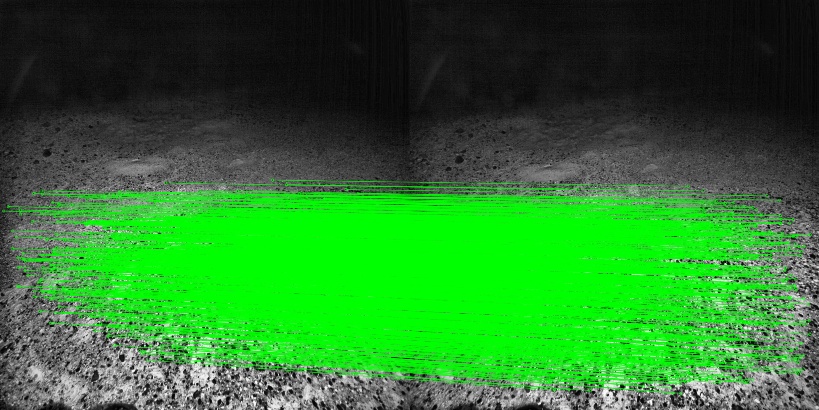}
    \caption{BEV: \SI{2.0}{Hz}}
  \end{subfigure}  
  \begin{subfigure}{0.23\textwidth}
    \includegraphics[width=\linewidth]{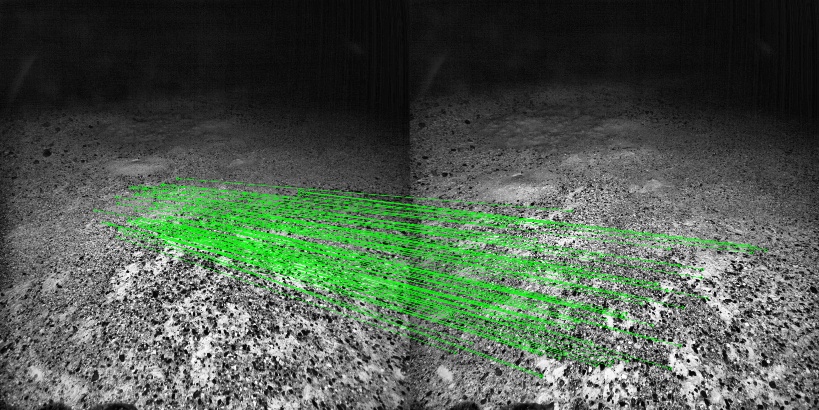}
    \caption{BEV: \SI{0.25}{Hz}}
  \end{subfigure}
   \par\medskip 
  \begin{subfigure}{0.23\textwidth}
    \includegraphics[width=\linewidth]{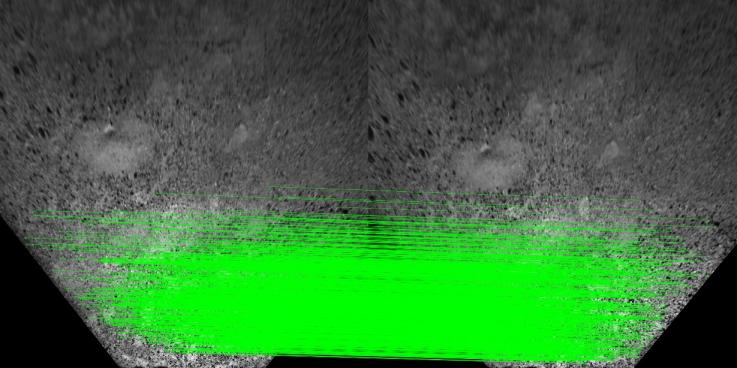}
    \caption{BEV$^{*}$: \SI{2.0}{Hz}}
  \end{subfigure}
  \begin{subfigure}{0.23\textwidth}
    \includegraphics[width=\linewidth]{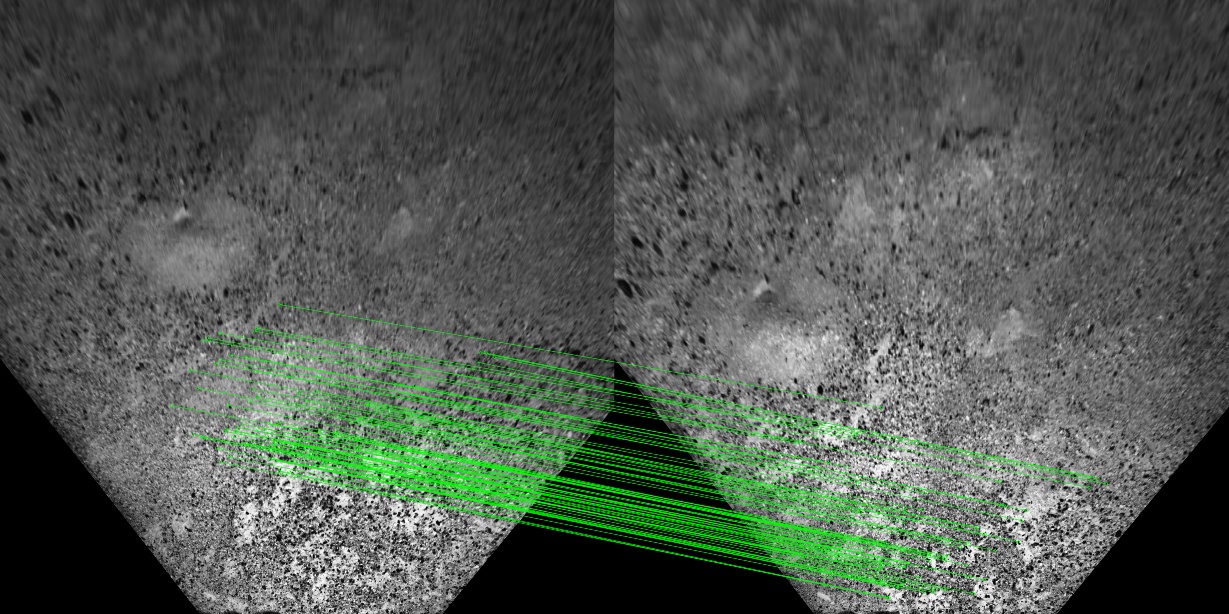}
    \caption{BEV$^{*}$: \SI{0.25}{Hz}}
  \end{subfigure}

  \caption{\small Nighttime feature matching comparison between feature matching in IS and BEV between images at \SI{2.0}{Hz} and \SI{0.25}{Hz}. Here, BEV denotes feature matched in BEV perspective and visualized in IS for comparability, and BEV$^{*}$ denotes that the features are matched in BEV and visualized in  BEV perspective.}
  \label{fig:night_ft_match}
  \vspace{-5px}
\end{figure}

\cref{fig:day_ft_match} shows the daytime case, where we observe that in the first row the matched features at \SI{2.0}{Hz} have large coverage over the image, whereas at \SI{0.25}{Hz} the number of features reduces and are distributed towards far-off points in the top region of the image. In the second row, we observe that \ac{bev} at \SI{0.25}{Hz} matches features much closer to the rover. The last row shows the matched feature in the \ac{bev} perspective, visualized in \ac{bev}. \cref{fig:night_ft_match} shows a similar comparison for night time, where at \SI{0.25}{Hz} \ac{is} matches far lesser features compared to \ac{bev} for the same \ac{fps}. Lastly, we see that at nighttime, \ac{bev} matches a higher number of features, closer to the rover.

\begin{figure}[]
    \centering
    \begin{subfigure}[b]{0.238\textwidth}
        \centering
        \includegraphics[width=\textwidth]{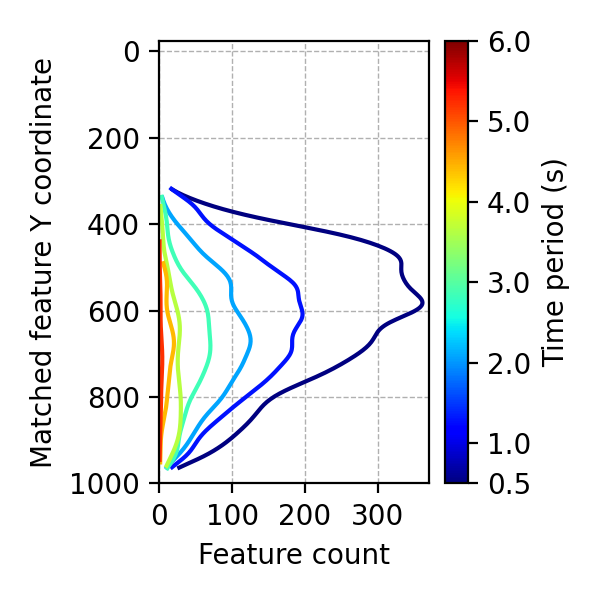}
        \caption{BEV: Daytime}
        \label{fig:ft_dist_bev_day_time}
    \end{subfigure}
    \hfill
    \begin{subfigure}[b]{0.238\textwidth}
        \centering
        \includegraphics[width=\textwidth]{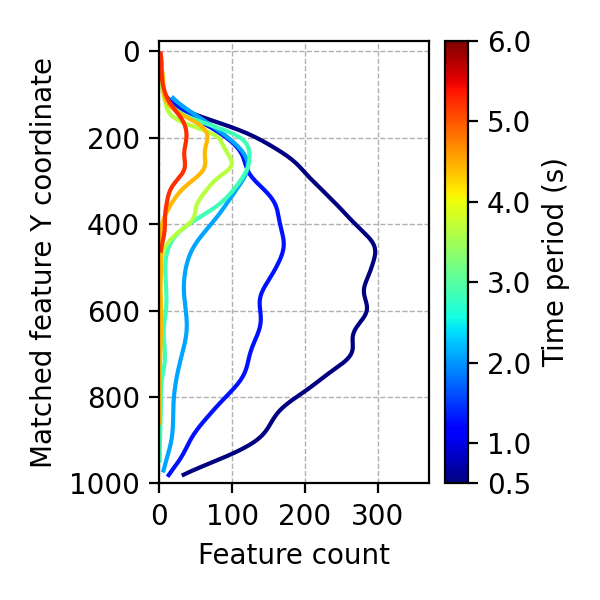}
        \caption{IS: Daytime}
        \label{fig:ft_dist_is_day_time}
    \end{subfigure}
    
    \vspace{1em} 
    
    \begin{subfigure}[b]{0.23\textwidth}
        \centering
        \includegraphics[width=\textwidth]{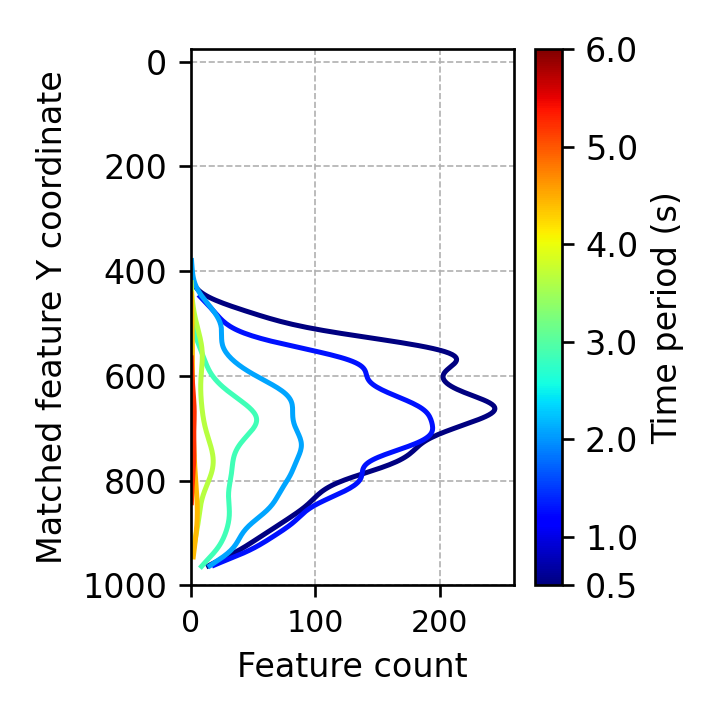}
        \caption{BEV: Nighttime}
        \label{fig:ft_dist_bev_night_time}
    \end{subfigure}
    \hfill
    \begin{subfigure}[b]{0.23\textwidth}
        \centering
        \includegraphics[width=\textwidth]{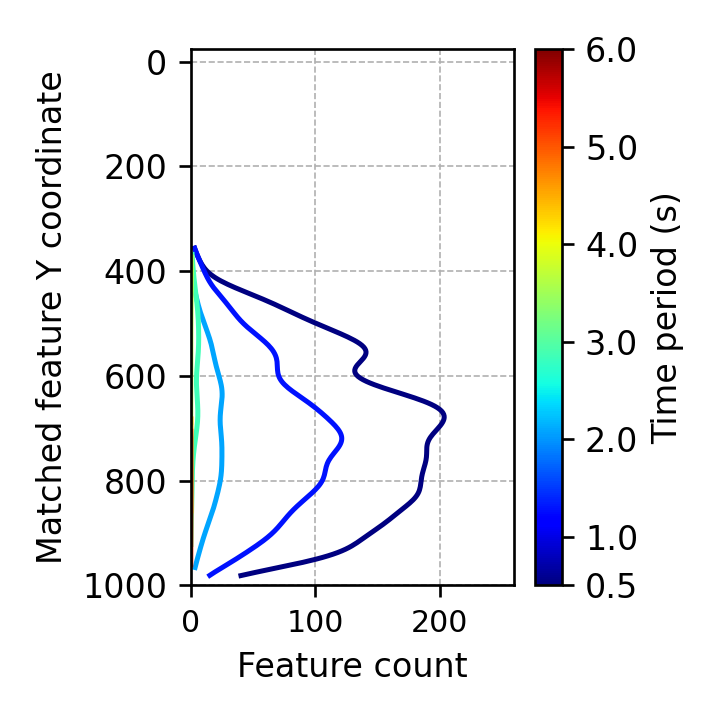}
        \caption{IS: Nighttime}
        \label{fig:ft_dist_is_night_time}
    \end{subfigure}
    \caption{\small Distribution of the feature locations along Y axis.}
    \label{fig:ft_distribution}
    \vspace{-5px}
\end{figure}

To assess the distributions of matched features at more intervals of time-periods between images, we overlay the smoothed histogram curves of the features along the $Y$ axis in the image coordinates to compare the distributions. This can be seen in \cref{fig:ft_distribution}. For features from nearby scenes (i.e., points in the lower region of the image), the distribution flattens much earlier in \ac{is} than in \ac{bev} for both day and nighttime scenes. Specifically, in the case of \ac{is}, the distributions in the lower region of the image flattens above \SI{2.0}{s} for both day and nighttime scenes, while \ac{bev} delays the flattening to over \SI{4.0}{s}. Showcasing its ability to match reliable features across larger baselines. For day-time (\cref{fig:ft_dist_is_day_time}), the \ac{is} distribution shifts to the upper region of the plot with increasing time-period, indicating that less informative far-off features are being matched. For nighttime scenes, (\cref{fig:ft_dist_bev_night_time}) \ac{bev} matches substantially more features, than \ac{is} (\cref{fig:ft_dist_is_night_time}). Lastly, due to limited self-illumination around the rover, far-off features are not matched in nighttime. This demonstrates the effectiveness of the \ac{bev} approach for matching visual features in planetary scenarios during both day and night.
\vspace{-2px}

\subsection{Analysis of VIO}
\begin{figure}
    \centering
    \includegraphics[width=0.80\linewidth]{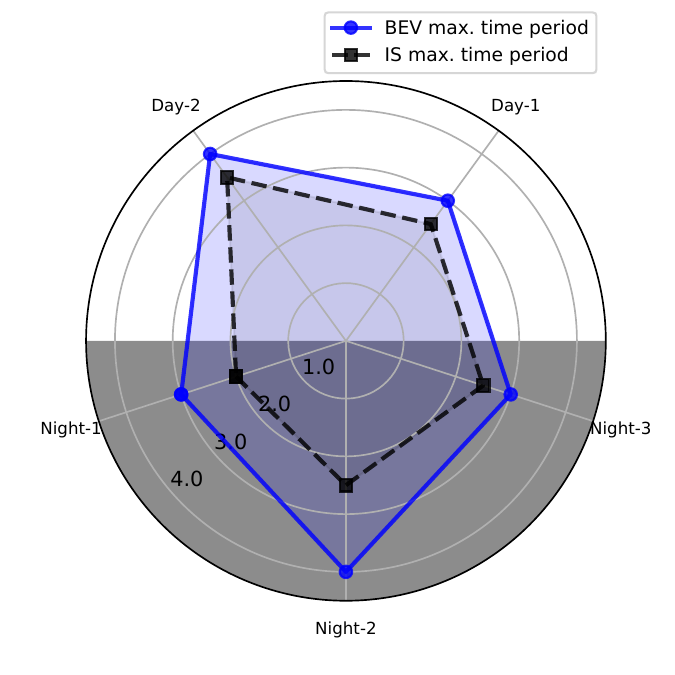}
    \caption{\small Envelope of maximum time period between image measurements where the VIO method runs successfully}
    \label{fig:envelope}
    \vspace{-11px}
\end{figure}

We present an ablation study on odometry performance across the five distinct trajectories, evaluating \ac{vio} performance with \ac{rpe} (given subsequent trajectory fragments lengths of \SI{10}{m}) against ground-truth from GPS. We progressively reduce the frame rate of images by skipping the images from the dataset in both day and nighttime conditions, using real and simulated data. The numerical analysis of the \ac{vio} for different scenarios is compiled in \cref{tab:vio_rpe}, which shows the \ac{rmse} of the \ac{rpe}. A common failure mode is a decrease in the number of matched features matched to as low as $0$, which causes large errors in the subsequent updates of the Kalman Filter, eventually causing the filter to diverge. We report the minimum frequency at which the \ac{vio} runs successfully in \cref{tab:envelope_table}. Based on this evaluation, we showcase the envelope of successful performance of the proposed \ac{bev} vs. the baseline \ac{is} in \cref{fig:envelope}.

\input{tex/main_results_table}

\begin{table}[t]
\centering
\caption{Minimum reliable image FPS for \ac{bev} and \ac{is}.}
\label{tab:envelope_table}
\begin{tabular}{lcccc}
\toprule
\textbf{Condition} & \textbf{BEV (Hz)} & \textbf{IS (Hz)} & \textbf{(\%)} \\
\midrule
Night-1        & 0.33 & 0.50 & 66.7 \\
Night-2        & 0.25 & 0.40 & 62.5 \\
Night-3        & 0.33 & 0.40 & 82.5 \\
Day-1          & 0.33 & 0.40 & 82.5 \\
Day-2          & 0.25 & 0.28 & 89.3 \\
\midrule
\textbf{Overall Mean} & & & \textbf{76.6} \\
\textbf{Nighttime Mean} & & & \textbf{70.3} \\
\textbf{Daytime Mean} & & & \textbf{85.9} \\
\bottomrule
\end{tabular}
\end{table}
\begin{figure}[t]
    \centering
    \includegraphics[width=0.48\textwidth]{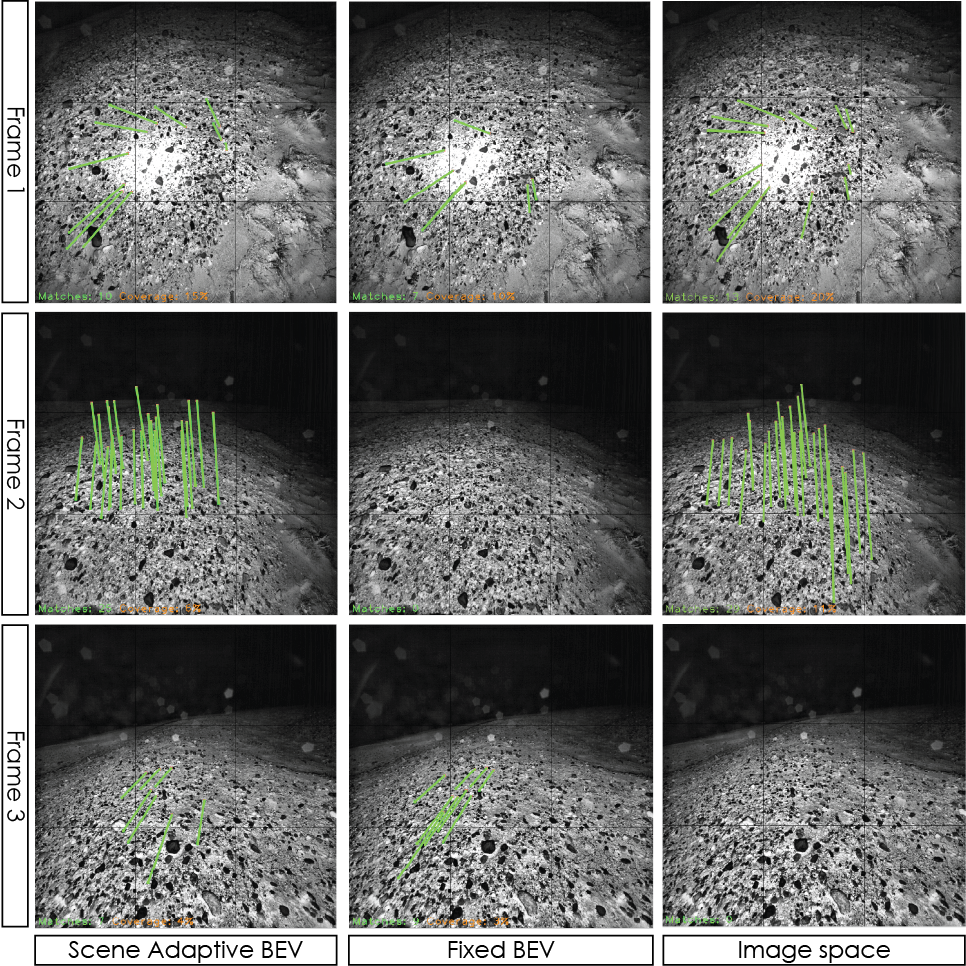}
    \caption{\small Comparison of feature matching performance over multiple subsequent images. The columns represent different configurations of the \ac{vio}, while the time increases from top to bottom.}
    \label{fig:3_comparision}
    \vspace{-12px}
\end{figure}

\subsection{Scene adaptive BEV, Fixed BEV, and IS}

We compare three cases of running \ac{vio}:
$(i)$ the proposed scene-adaptive homography estimation for \ac{bev},
$(ii)$ a fixed ideal homography for \ac{bev}, and
$(iii)$ the baseline using \ac{is}.
\Cref{fig:3_comparision} presents the qualitative results for each case.

The experiments were conducted at an image update frequency of \SI{0.33}{Hz}, with the rover first traversing uneven terrain that induced large variations in attitude, followed by a flat region. The attitude variations are visible between the first and second rows of images, while the transition to flat terrain occurs between the second and third rows.

The proposed scene-adaptive \ac{bev} (first column) consistently maintains reliable feature matches across both conditions. In contrast, case $(ii)$ struggles under large attitude changes, while case $(iii)$ fails to maintain matches in the flat region due to significant perspective changes.

%% file: tex/main_results_table.tex

\begin{table}[t]
\centering
\tiny
\caption{Comparison of \ac{rmse} of \ac{rpe} in meters for \ac{vio} with \ac{bev} and \ac{is}}
\label{tab:vio_rpe}
\begin{tabular}{llcccccccc}
\toprule
\multicolumn{2}{c}{\textbf{Time-period (s)}} & 0.5 & 1.0 & 2.0 & 2.5 & 3.0 & 3.5 & 4.0 \\ 
\cmidrule(lr){3-9}
\multicolumn{2}{c}{\textbf{Frequency (Hz)}} & 2.0 & 1.0 & 0.5 & 0.4 & 0.33 & 0.28 & 0.25 \\ 
\midrule
\multirow{2}{*}{Night-1 (real-world)} & BEV      & 0.263 & 0.273 & 0.271 & 0.276 & 0.341 &   -    &   -    \\
                         & \ac{is} & 0.282 & 0.274 & 0.274 &   -    &   -    &   -    &   -    \\
\midrule
\multirow{2}{*}{Night-2 (real-world)} & BEV      & 0.129 & 0.127 & 0.123 & 0.122 & 0.126 & 0.126 & 0.213 \\
                         & \ac{is} & 0.124 & 0.124 & 0.126 & 0.126 &   -    &   -    &   -    \\
\midrule
\multirow{2}{*}{Night-3 (simulated)} & BEV      & 0.281 & 0.300 & 0.304 & 0.303 & 0.310 &   -    &   -    \\
                         & \ac{is} & 0.285 & 0.309 & 0.299 & 0.370 &   -    &   -    &   -    \\
\midrule
\multirow{2}{*}{Day-1 (real-world)}   & BEV      & 0.282 & 0.268 & 0.252 & 0.273 & 0.266 &   -    &   -    \\
                         & \ac{is} & 0.322 & 0.261 & 0.257 & 0.240 &   -    &   -    &   -    \\
\midrule
\multirow{2}{*}{Day-2 (real-world)}   & BEV      & 0.172 & 0.172 & 0.172 & 0.172 & 0.178 & 0.201 & 0.230 \\
                         & \ac{is} & 0.175 & 0.174 & 0.175 & 0.179 & 0.188 & 0.199 &   -    \\
\bottomrule
\end{tabular}
\vspace{-12pt}
\end{table}

%% file: tex/6_Conclusion.tex
In this work, we tackle the problem of visual-inertial state estimation for a lunar rover, in the context of the Endurance mission concept. Such a system faces challenges due to constrained computational and energy resources; it is tasked to navigate across day and nighttime scenes, operating under solar or self-illumination. We proposed a solution based on scene adaptive \ac{bev} that enables increasing the temporal and spatial sparsity of the image updates (i.e., the image \ac{fps}). The perspective equalizing property of BEV explains its ability to maintain reliable matches at larger inter-frame baselines, enabling sparser visual updates. The rover navigation benefits from increased sparsity in the visual updates, both by lowering computational and energy requirements associated with computing, camera exposure, and strobing LEDs in dark scenes. We compare our approach against a baseline \ac{vio} method that uses feature matching in \ac{is}. In our quantitative evaluations from data collected across day and night on a half-scale lunar rover prototype in a planetary-like environment, we demonstrate that the proposed method enables \ac{vio} at a minimum update rate of \SI{0.25}{Hz}, which is $62.5\%$ of the minimum update rate of the baseline method. Overall, we demonstrate that our approach operates in a larger envelope of sparse visual updates.

Future work may focus on developing a deeper understanding of the proposed \ac{bev} perspective and its influence on \ac{vio}. In particular, it is necessary to investigate whether feature matching in \ac{bev} alters the pixel measurement noise characteristics, and to determine if commonly assumed constant noise model remains valid. Furthermore, the impact of feature matching on objects that deviate from the estimated plane fit in the 3D point cloud warrants further investigation.